\documentclass[runningheads,a4paper]{llncs}

\usepackage{amssymb}
\usepackage{amsmath,graphicx}
\usepackage{amssymb}
\usepackage{algorithm}
\usepackage{algpseudocode}
\usepackage{multirow}
\usepackage{array}
\usepackage{url}
\usepackage{color}

\begin{document}

\mainmatter  

\title{Tool and Phase recognition using contextual CNN features}
\titlerunning{Surgical Tool and Phase recognition using contextual CNN features}

\author{Manish Sahu, Anirban Mukhopadhyay, Angelika Szengel \and Stefan Zachow}

\institute{Zuse Institute Berlin, Berlin, Germany}

%


%
%

\maketitle

\begin{abstract}

A transfer learning method for generating features suitable for surgical tools and phase recognition from the ImageNet classification features~\cite{Krizhevsky12} is proposed here.  
In addition, methods are developed for generating contextual features and combining them with time series analysis for final classification using multi-class random forest.
The proposed pipeline is tested over the training and testing datasets of M2CAI16 challenges: tool and phase detection.
Encouraging results are obtained by leave-one-out cross validation evaluation on the training dataset.

\end{abstract}

\section{Introduction}  \label{sec:intro}

Fully automatic recognition of different phases in surgical workflow is a core unresolved task of computer assisted intervention (CAI). 
Various applications e.g. context aware surgical systems, staff assignment, automated guidance during intervention, surgical alert systems etc. can benefit from fully automatic recognition of surgical phases.

Recently, EndoNet~\cite{Twinanda16} has been proposed, which learns features suitable for surgical workflow detection and estimation, by transferring the CNN features learnt for ImageNet classification task~\cite{Krizhevsky12}. 
In particular, EndoNet has achieved significant success in detecting and recognizing surgical tools during a CAI, which in turn is shown to be useful in recognizing phases of surgical workflow.         
However, we have identified some key areas e.g. employing context, hard negative mining etc. that can be employed in association with the transfer learning technique described in EndoNet~\cite{Twinanda16}. This report consists of preliminary investigations along those directions.

\section{Method}  \label{sec:method}

This section describes the method proposed for surgical tool and phase recognition.

\subsection{Transfer Learning ImageNet Features} \label{subsec:CNN}

In recent years CNNs have been used extensively for detection, localization and segmentation tasks, however, training a CNN from scratch is not a common practice when the data-set is not large enough. 
Instead the pre-trained convolutional networks are used either as a fixed feature detector or the weights are used to initialize the similar network followed by fine tuning the network for a specified task.

\begin{figure}[t!]
  \centering
  \centerline{\includegraphics[width=12cm]{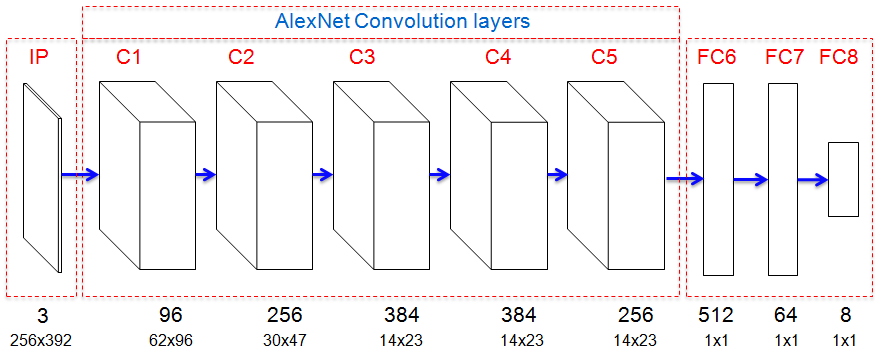}}
  \caption{Proposed CNN architecture. The convolution layers are same as AlexNet~\cite{Krizhevsky12} architecture. }
\label{fig:CNN_training}
\end{figure}

For the tool and phase detection tasks, we have created the CNN architecture (see Fig. \ref{fig:CNN_training}) which consists of convolutional layers similar to AlexNet architecture but input and fully connected layers specific to tool detection task. 
On top of the convolutional layers fully connected layers: 'fc6', 'fc7' and 'fc8' of size 512, 64 and 8 respectively. 
Moreover, we have added an additional label 'No Tool' which means that none of the tool is visible in the endoscopic image. 
The output layer is connected to 'fc8' layer having 8 units (seven tools + 'No tool' label). 
Since the convolutional filters of the network are independent of the input dimension, its weights are initialized with the AlexNet pre-trained weights. 
The learning is formulated as a multi-classification problem with cross entropy as the loss function. 
The network is trained using stochastic gradient descent (learning rate of 0.1) with momentum of 0.9 until convergence.
During learning, random cropping and flipping is performed for artificial data augmentation.
For phase detection, contextual features are generated by concatenating 'fc7' features from ten previous time points.

\subsection{Time Series} \label{subsec:TS}

The inherent ordering and mutual exclusiveness of surgical phases can be exploited further by exploiting time series techniques. 
Our key observation in this regard is that, appearance based contextual CNN features described in Section~\ref{subsec:CNN}, results in misclassification between phases far away in the temporal ordering. We have addressed this issue by modeling the temporal occurrences of surgical phases by fitting Gaussian distribution. 
Though it'd be ideal to fit each phase with a Gaussian distribution, due to confusing temporal occurrences of different phases, we have considered only three Gaussian distributions. 
First one fits the two initial phases, second one fits the three middle phases and the last three phases are fitted by the third distribution.

\subsection{Random Forest classification} \label{subsec:RF}  

Random Forest is used in a multi-class classification setting for both challenges. 
For tool recognition challenge, we have simply fed the CNN features in a random forest classifier during training, and used the trained classifier for prediction during testing. 

For surgical phase recognition challenge, we have used a two step classification approach. 
The first step is simple multi-class classification using random forest. 
The prediction probability of this first stage classifier is combined to Time series to have a more localized prediction. 
In a follow-up~\emph{hard negative mining} step, we have trained eight random forest classifier, one for each phase, where only the neighboring phases are considered for training. 
The features corresponding to the joint predicted labels of time series and initial phases are fed to the corresponding random forest for the final prediction of surgical phase.

\section{Results}  \label{sec:results}

This section describes the results of our proposed method on M2CAI16 challenge training dataset. All the experiments are performed in a strict leave-one-video-out cross validation setup. 

\subsection{Tool Recognition}

Result of tool recognition is presented in Table~\ref{table:tool_table}. In particular, Average Precision (AP) for each tool is reported. 
It is interesting that AP of Scissors is lower than the rest of the tools. 
This result is similar to that of Endonet~\cite{Twinanda16}. However, a detailed investigation is needed for further understanding. 
In testing dataset, mean AP for our proposed method is \textbf{61.5}.

\begin{table}[t] \label{table:tool_table}
	\centering
	\caption{Average Precision (AP) for all tools, computed on the M2CAI16 tool detection challenge training dataset.}
	\begin{tabular}{|l|c|}
		\hline
		Tool   & AP \\
		\hline 
		\hline
		Bipolar & 40.8 \\ 
		Clipper & 35.3 \\ 
		Grasper & 73.9 \\ 
		Hook & 95.1 \\ 
		Irrigator & 33.2 \\ 
		Scissors & 26.2 \\ 
		Specimen Bag & 76.6 \\ 
		\hline
		\textbf{Mean}   & \textbf{54.5} \\
		\hline
	\end{tabular}
	
\end{table}

\subsection{Phase Recognition}

In Table~\ref{table:phase_table}, F1-score of phase recognition is reported. 
In particular, the Preparation phase has the lowest F1-score. 
The mean F1-score of our proposed method over all surgical phases is \textbf{53.13}. 

\begin{table}[t] \label{table:phase_table}
	\centering
	\caption{Accuracy for all surgical phases, computed on the M2CAI16 phase detection challenge training dataset.}
	\begin{tabular}{|l|c|}
		\hline
		Tool   & F1-Score \\
		\hline 
		\hline
		Trocar Placement & 78.07  \\ 
		Preparation & 39.59 \\ 
		Calot Triangle Dissection & 56.11  \\ 
		Clipping and Cutting & 54.18 \\ 
		Gallbladder Dissection & 47.96 \\ 
		Gallbladder Packaging & 51.78 \\ 
		Cleaning Coagulation & 46.36 \\
		Gallbladder Retraction & 51.00 \\ 
		\hline
		\textbf{Average}   & \textbf{53.13} \\
		\hline
	\end{tabular}
	
\end{table}

\section{Discussions and Conclusion} \label{sec:disc}

In this report, we have addressed the problem of surgical tool and phase recognition. 
Based on some key observations, transfer learnt ImageNet classification features are combined to create contextual features. 
Furthermore, time series and hard negative mining is combined for classifying phases using Random Forest.
Resulting classification accuracy in tool and phase recognition is promising considering the size of the training data. 
One particular limitation of the proposed technique is the rudimentary concatenation of features for contextual information generation. 
In future, we are planning explore the possibilities of more involved feature fusion techniques for improving the contextual features.

\end{document}